%% file: main.tex
\documentclass[runningheads]{llncs}

\usepackage[mobile]{eccv}

\usepackage{eccvabbrv}
\usepackage{amssymb}
\usepackage{graphicx}
\usepackage{booktabs}
\usepackage{multirow}
\usepackage{xcolor}

\usepackage[accsupp]{axessibility}  

\usepackage{wrapfig}

\usepackage[pagebackref,breaklinks,colorlinks,citecolor=blue]{hyperref}

\usepackage{orcidlink}

\newcommand{\repeatthanks}{\textsuperscript{\thefootnote}}

\begin{document}

\title{\raisebox{-0.18cm}{\includegraphics[scale=0.18 ]{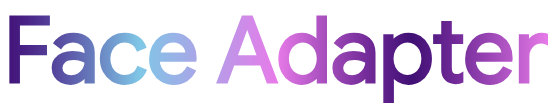}} \\ for Pre-Trained Diffusion Models \\ with Fine-Grained ID and Attribute Control} 

\titlerunning{Face-Adapter}

\author{Yue Han\thanks{co-first authors; $^{\dagger}$corresponding author} \inst{1} \and
Junwei Zhu\repeatthanks \inst{2}\and
Keke He\inst{2}\and
Xu Chen\inst{2}\and
Yanhao Ge\inst{3}\and
Wei Li\inst{3}\and
Xiangtai Li\inst{4}\and
Jiangning Zhang\inst{2}\and
Chengjie Wang\inst{2},
Yong Liu$^{\dagger,1}$
}

\authorrunning{Y. Han et al.}


\institute{$^{1}$Zhejiang University, $^{2}$Tencent, $^{3}$VIVO, $^{4}$Nanyang Technological University  
\email{12432015@zju.edu.cn, yongliu@iipc.zju.edu.cn} \\
\url{https://faceadapter.github.io/face-adapter.github.io/}}

\maketitle
\input{sections/0_abstract}

\input{sections/1_intro}

\input{sections/2_related}

\input{sections/3_method}

\input{sections/4_experiments}

\input{sections/5_conclusion}

%
%


\bibliographystyle{splncs04}
\bibliography{main}

\end{document}

%% file: sections/0_abstract.tex
\begin{abstract}
Current face reenactment and swapping methods mainly rely on GAN frameworks, but recent focus has shifted to pre-trained diffusion models for their superior generation capabilities. However, training these models is resource-intensive, and the results have not yet achieved satisfactory performance levels.
To address this issue, we introduce \textbf{Face-Adapter}, an efficient and effective adapter designed for high-precision and high-fidelity face editing for pre-trained diffusion models. 
We observe that both face reenactment/swapping tasks essentially involve combinations of target structure, ID and attribute. We aim to sufficiently decouple the control of these factors to achieve both tasks in one model. 
Specifically, our method contains: 
1) A Spatial Condition Generator that provides precise landmarks and background;
2) A Plug-and-play Identity Encoder that transfers face embeddings to the text space by a transformer decoder. 
3) An Attribute Controller that integrates spatial conditions and detailed attributes. 
Face-Adapter achieves comparable or even superior performance in terms of motion control precision, ID retention capability, and generation quality compared to fully fine-tuned face reenactment/swapping models. 
Additionally, Face-Adapter seamlessly integrates with various StableDiffusion models.

\end{abstract}
\keywords{Face Reenactment \and Face Swapping \and Diffusion Model}

%% file: sections/1_intro.tex
\section{Introduction}
\label{sec:intro}
\begin{figure*}[tp]
    \centering
    \includegraphics[width=0.90\linewidth]{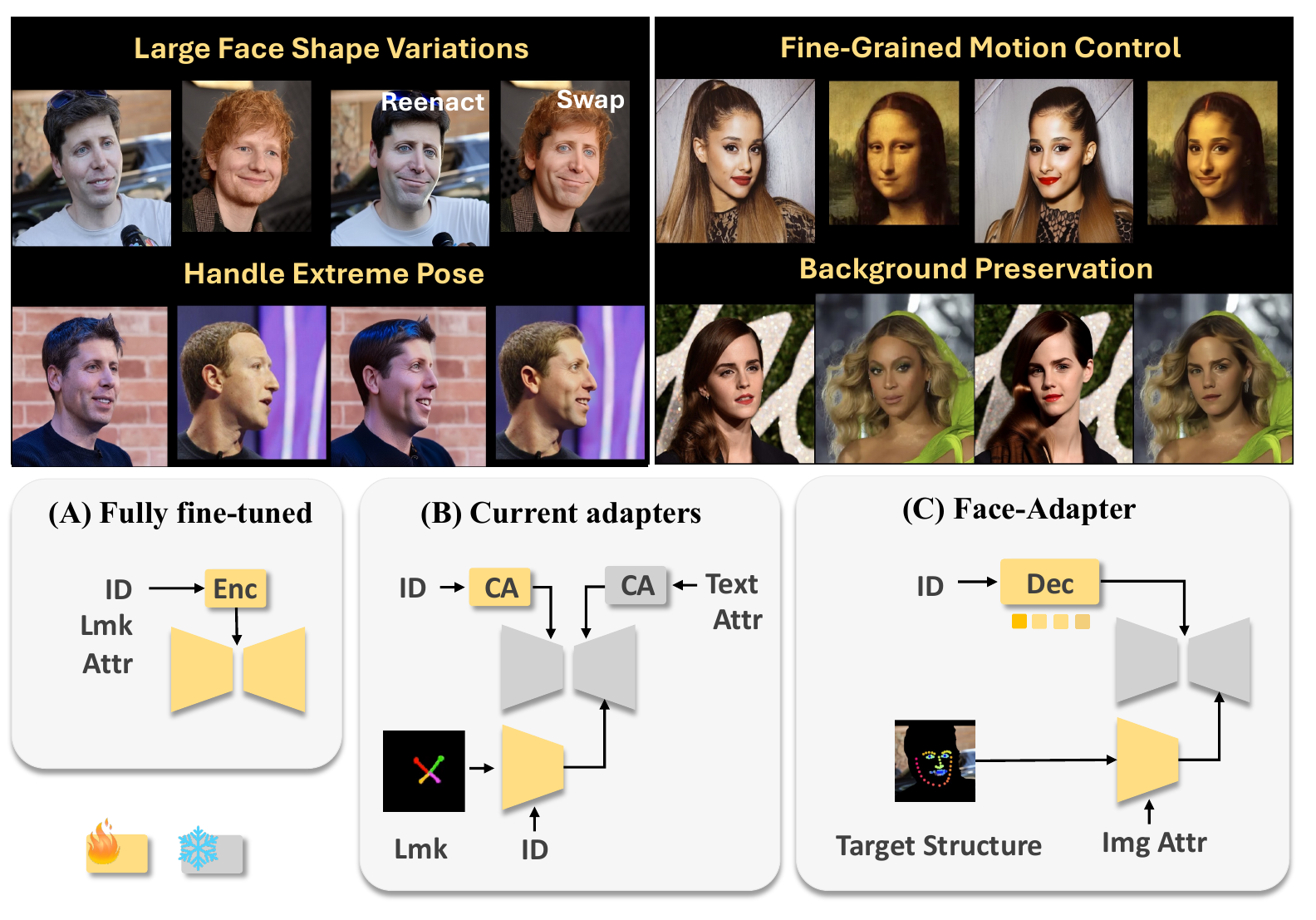}
    \vspace{-1.0em}
    \caption{   
    \textbf{Top:}  Face-Adapter supports a 'one-model-two-tasks' approach and demonstrates robustness under various challenging scenarios.
    \textbf{Bottom:} The design motivation is (1) Both face reenactment and swapping require fully disentangled ID, target structure, 
    and attribute control; (2) Addressing overlooked issues unified in target structure; (3) Effective ID injection avoids SD fine-tuning, making Face-Adapter plug-and-play. 
    }
    \label{fig:teaser}
    
\end{figure*}

Face reenactment aims to transfer the target motion onto the source identity and attributes, while face swapping aims to transfer the source identity onto the target motion and attributes. Both tasks require complete disentangling and fine-grained control of identity, attributes, and motion. Current face reenactment and swapping techniques mainly rely on GAN-based frameworks~\cite{fomm,pirenderer,dg,tpsm,dam,hyperreenact,hififace,simswap,infoswap,blendface}. 
However, GAN-based methods encounter limitations in their generative capabilities, making it challenging to tackle hard cases, such as handling large poses in face reenactment and accommodating facial shape variations in face swapping.

Existing studies~\cite{fadm, diffswap} have attempted to address these challenges by leveraging the powerful generative capabilities of the diffusion models. However, these methods necessitate full model training, resulting in significant computational overhead, and they have not been successful in delivering satisfactory outcomes. For instance, FADM~\cite{fadm} refines the results of GAN-based reenactment methods, which improves image quality but still fails to resolve the blurring issue caused by large pose variation. 
On the other hand, DiffSwap~\cite{diffswap} produces blurry facial outcomes due to the lack of background information during training, which hampers model learning. 
Moreover, these methods do not fully exploit the potential of large pre-trained diffusion models.
To reduce training costs, some methods~\cite{ipadapter, instantid} have introduced face editing adapter plugins for large pre-trained diffusion models. 
However, these approaches primarily focus on attribute editing using text, which inevitably weakens spatial control to ensure text editability. 
For example, they can only use five points~\cite{instantid} to control facial poses, limiting their ability to control expressions and gaze precisely.
On the other hand, direct inpainting with masks of the face area does not take into account facial shape changes, leading to a decrease in identity preservation.

To address the above challenges, we are committed to developing an efficient and effective face editing adapter (\textbf{Face-Adapter}) for pre-trained diffusion models, specifically targeting face reenactment and swapping tasks. 
The design motivation of Face-Adapter is threefold: (1) Fully disentangled ID, target structure, and attribute control enable a 'one-model-two-tasks' approach; (2) Addressing overlooked issues; (3) Simple yet effective, plug and play. 
Specifically, the proposed Face-Adapter comprises three components: 
\textbf{1)} Spatial Condition Generator (SCG in \cref{sec:scg}) is designed to automatically predict 3D prior landmarks and the mask of the varying foreground area, which provides more reasonable and precise guidance for subsequent controlled generation. 
In addition, for face reenactment, this strategy mitigates potential problems that could occur when only extracting the background from the source image, such as inconsistencies caused by alterations in the target background due to the movement of the camera or face objects; 
For face swapping, the model learns to maintain background consistency, glean clues about global lighting and spatial reference, and try to generate content in harmony with the background.
\textbf{2)} Identity Encoder (IE in \cref{sec:ie}) uses the pre-trained recognition model to extract face embeddings and then transfers them to the text space by learnable queries from the transformer decoder. This manner greatly improves the identity consistency of the generated images. 
\textbf{3)} Attribute Controller (AC in \cref{sec:ac}) includes two sub-modules: The spatial control combines the landmarks of target motion with the unchanged background obtained from the Spatial Condition Generator. 
The attribute template supplements the absent attribute, encompassing lighting, a portion of the background, and hair. 
Both two tasks can be perceived as a procedure that executes conditional inpainting, utilizing the provided identity and absent attribute content. This process adheres to the stipulations of the given spatial control, attaining congruity and harmony with the background.
Our contributions can be summarized as follows:
\begin{itemize}
  \item We introduce Face-Adapter, a lightweight facial editing adapter designed to facilitate precise control over identity and attributes for pre-trained diffusion models. This adapter efficiently and proficiently tackles face reenactment and swapping tasks, surpassing previous state-of-the-art GAN-based and diffusion-based methods.
  \item We propose a novel Spatial Condition Generator module to predict the requisite generation areas, collaborating with the Identity Encoder and Attribute Controller to frame reenactment and swapping tasks as conditional inpainting with sufficient spatial guidance, identity, and essential attributes. Through reasonable and highly decoupled condition designs, we unleash the generative capabilities of pre-trained diffusion models for both tasks.
  \item Face-Adapter serves as a training-efficient, plug-and-play, face-specific adapter for pre-trained diffusion models. By freezing all parameters in the denoising U-Net, our method effectively capitalizes on priors and prevents overfitting. 
  Furthermore, Face-Adapter supports a "one model for two tasks" approach, enabling simple input modifications to independently accomplish superior or competitive results of two facial tasks on VoxCeleb1/2 datasets. 
\end{itemize}

%% file: sections/2_related.tex
\section{Related Work}
\label{sec:2}

\vspace{1mm}
\noindent\textbf{Face Reeactment} involves extracting motion from a human face and transferring it to another face~\cite{monkeynet, zeng2020realistic, freenet, xu2022designing, nirkin2019fsgan, agarwal2023audio, yang2022face2face,bounareli2023hyperreenact,zhang2023metaportrait}, which can be broadly divided into warping-based and 3DMM-based methods. 
\textit{Warping-based methods}~\cite{monkeynet,tpsm,fomm,face-vid2vid,dagan,mcnet} typically extract landmarks or region pairs to estimate motion fields and perform warping on the feature maps to transfer motions. 
%
%
When dealing with large motion variations, these methods tend to produce blurry and distorted results due to the difficulty in predicting accurate motion fields.
%
%
%
\textit{3DMM-based methods}~\cite{pirenderer} use facial reconstruction coefficients or rendered images from 3DMM as motion control conditions. 
The facial prior provided by 3DMM enables these methods to obtain more robust generation results in large pose scenarios. 
%
Despite offering accurate structure references, it only provides coarse facial texture and lacks references for hair, teeth, and eye movement.
%
%
StyleHEAT~\cite{styleheat} and HyperReenact~\cite{hyperreenact} use StyleGAN2 to improve generation quality. However, StyleHEAT is limited by the dataset of frontal portraits, while HyperReenact suffers from resolution constraints and background blurring.
To further improve generation quality, diffusion models have gained popularity.  
FADM~\cite{fadm} combines the previous reenactment model with diffusion refinements but the base model limits the driving accuracy.
Recently, AnimateAnyone~\cite{animateanyone} employs heavy texture representation encoders (CLIP and a copy of U-Net) to ensure the textural quality of animated results, but this manner is costly.
In contrast, we aim to leverage the generative capabilities of pre-trained T2I diffusion models fully and seek to comprehensively overcome the challenges presented in previous methods, \eg, low -resolution generation, difficulty in handling large variations, efficient training, and unexpected artifacts.

%
\vspace{1mm}
\noindent\textbf{Face Swapping} aims to transfer the facial identity of the source image to the target image, with other attributes (\ie, lighting, hair, background, and motion) of the target image unchanged. 
Recent methods can be broadly classified into GAN-based and diffusion-based approaches.
\textbf{1)} Most GAN-based methods~\cite{e4s,faceshifter,simswap, region,megafs,mobilefaceswap} are dedicated to resolving the disentanglement and fusion of the identity and other attributes. 
Efforts include introducing face parsing masks, various losses for attribute-preserving, and designing fusion modules. Despite promising improvement, these methods often produce noticeable artifacts when dealing with significant changes in face shape or occlusions. 
HifiFace~\cite{hififace} alleviates this issue by utilizing 3DMM to reconstruct a reference face which combines the source face shape with other attributes of the target. 
However, relying on GAN to ensure generation quality, HifiFace still fails to inpaint harmonious results when dealing with large blank areas caused by face shape variation. 
\textbf{2)} Diffusion-based methods utilize the generative capabilities of the diffusion model to enhance sample quality. However, the numerous denoising steps during inference significantly increase the training costs when using attribute-preserving loss. 
DiffSwap~\cite{diffswap} proposes midpoint estimation to address this issue, but the resulting error and the lack of background information for inpainting reference lead to unnatural results. Moreover, these methods require costly training from scratch. 
In contrast, our Face-Adapter ensure image quality only relying on the denoise loss with complete disentanglement of the control of the target structure, ID and other attributes.
Moreover, Face-Adapter further significantly reduces training costs by freezing all of U-Net's parameters, which also preserves prior knowledge and prevents overfitting.

\vspace{1mm}
\noindent\textbf{Personalization of Pretrained Diffusion Models.} 
Personalization aims to insert a given identity into the pre-trained T2I diffusion models. 
Early works~\cite{textual_inversion,dreambooth} insert identity by using optimization or fine-tuning manners. 
Subsequent studies~\cite{custom,fastcomposer,portraitbooth} introduce coarse spatial control, achieving multi-subject generation and regional attribute editing with text, but these methods require fine-tuning of most parameters.
IP-adapter(-FaceID)~\cite{ipadapter} and InstantID~\cite{instantid} fine-tune only a few parameters. The latter achieves robust identity preservation. However, as a tradeoff for text editability, InstantID could only apply weak spatial control. Therefore, it struggles with fine movements (expression and gaze) in face reenactment and swapping. 
By comparison, our Face-Adapter is an effective and lightweight adapter designed for pre-trained diffusion models to accomplish face reenactment and swapping simultaneously.

%% file: sections/3_method.tex







\section{Methods}
\label{sec:methods}

\begin{figure*}[tp]
    \centering
    \includegraphics[width=0.97\linewidth]{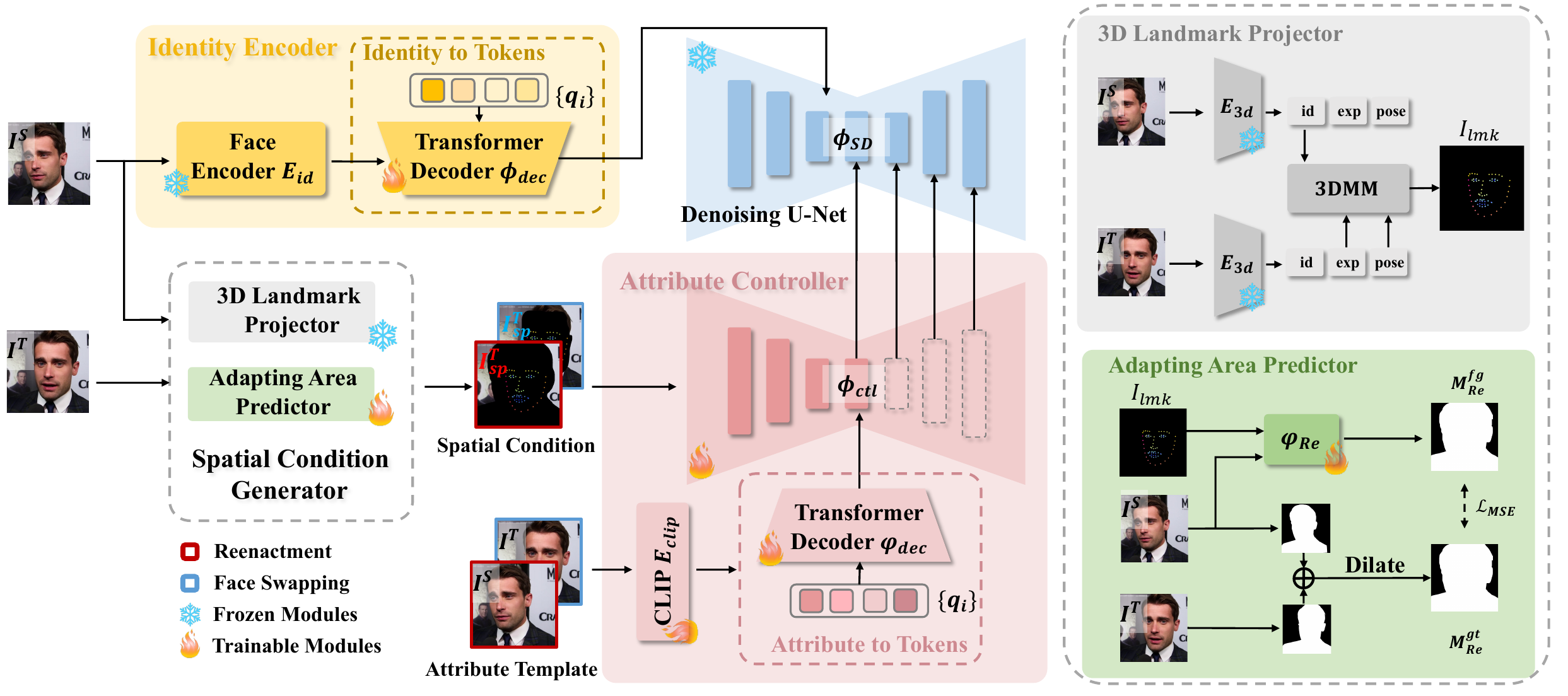}
    \caption{\textbf{Overview pipeline of our proposed Face-Adapter} that consists of three modules: 
    %
    %
    1) The Spatial Condition Generator predicts 3D prior landmarks and adapts the foreground mask automatically, offering more accurate guidance for controlled generation. %
    2) The Identity Encoder improves identity consistency in generated images by transferring face embeddings to the text space using learnable queries. %
    3) The Attribute Controller features (i) spatial control that combines target motion landmarks with the invariant background from the Spatial Condition Generator, and (ii) an attribute template to fill in missing attributes.
    }
    \label{fig:pipeline}
\end{figure*}




The comprehensive structure of the proposed Face-Adapter is illustrated in~\cref{fig:pipeline}, which aims to integrate identity into the attribute template, which provides essential attributes (\eg, lighting, a portion of the background, and hair) based on the target motion (\eg, pose, expression, and gaze). 

\subsection{Spatial Condition Generator}
\label{sec:scg}
To provide more reasonable and precise guidance for subsequent controlled generation, we design a novel Spatial Condition Generator (SCG) to automatically predict 3D prior landmarks and the mask of the varying foreground area. In detail, this component consists of two sub-modules: 

\noindent\textbf{3D Landmark Projector.} 
To surmount alterations in facial shape, we utilize a 3D facial reconstruction method~\cite{d3dfr} to extract the identity, expression individually and pose coefficients of the source and target faces. Subsequently, we recombine the identity coefficients of the source with the expression and pose coefficients of the target, reconstruct a new 3D face, and project it to acquire the corresponding landmarks.

\begin{figure*}[h!]
    \centering
    \includegraphics[width=0.95\linewidth]{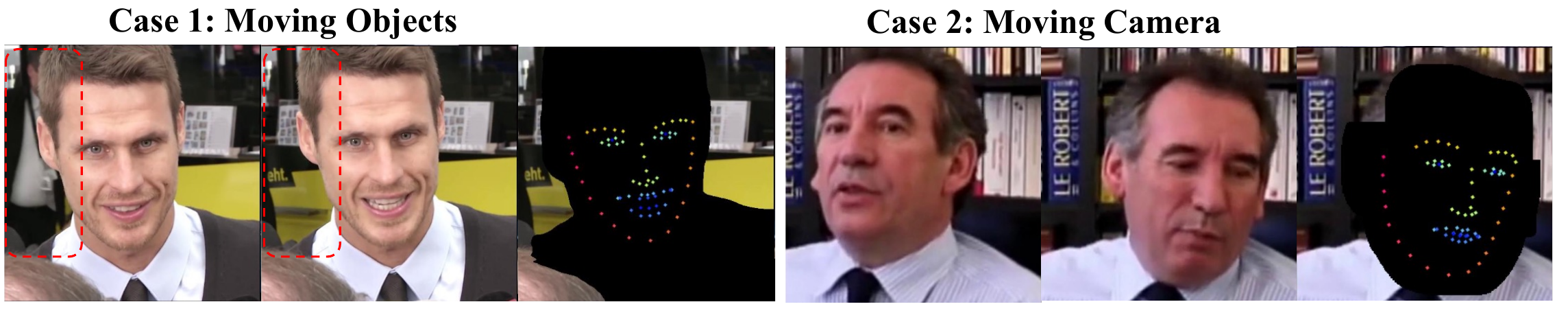}
    \caption{
    Background inconsistency between the input (\ie, source) and the groundtruth (\ie, target) makes the model confused and fail to learn to generate clear background. Thus, we provide the background of the target image in the spatial condition during training to address this inconsistency.
    }
    \label{fig:background_movement}
\end{figure*}

\noindent\textbf{Adapting Area Predictor.} 
For face reenactment, prior methods assume that only the subject is in motion, while the background remains static in the training data. However, we observe that the background actually undergoes changes, encompassing the movement of both the camera and objects in the background, as illustrated in~\cref{fig:background_movement}. If the model lacks knowledge of the background motion during training, it will learn to generate a blurry background.
For face swapping, supplying the target background can also give the model clues about environmental lighting and spatial references. 
This added constraint of the background significantly diminishes the difficulty of the model learning, transitioning it from learning a task of generating from scratch to a task of conditional inpainting. As a result, the model becomes more attuned to preserving background consistency and generating content that seamlessly integrates with it. 

In view of the above, we introduce a lightweight Adapting Area Predictor for both face reenactment and swapping, automatically predicting the region the model needs to generate (the adapting area) while maintaining the remaining area unchanged. 
For face reenactment, the adapting area constitutes the region occupied by the source image head before and after reenactment. We train a mask predictor $\varphi_{Re}$ that accepts the target image $I^T$ and motion landmarks $I_{lmk}$ from the 3D Landmark Projector to predict the adapting area mask $M^{fg}_{Re}$. 
The mask ground truth $M^{gt}_{Re}$ is generated by taking the union of the head regions (including hair, face, and neck) of the source and target, followed by outward dilation. Head regions are obtained using a pre-trained face parsing model~\cite{bisenet}.
\begin{figure*}[tp]
    \centering
    \includegraphics[width=0.8\linewidth]{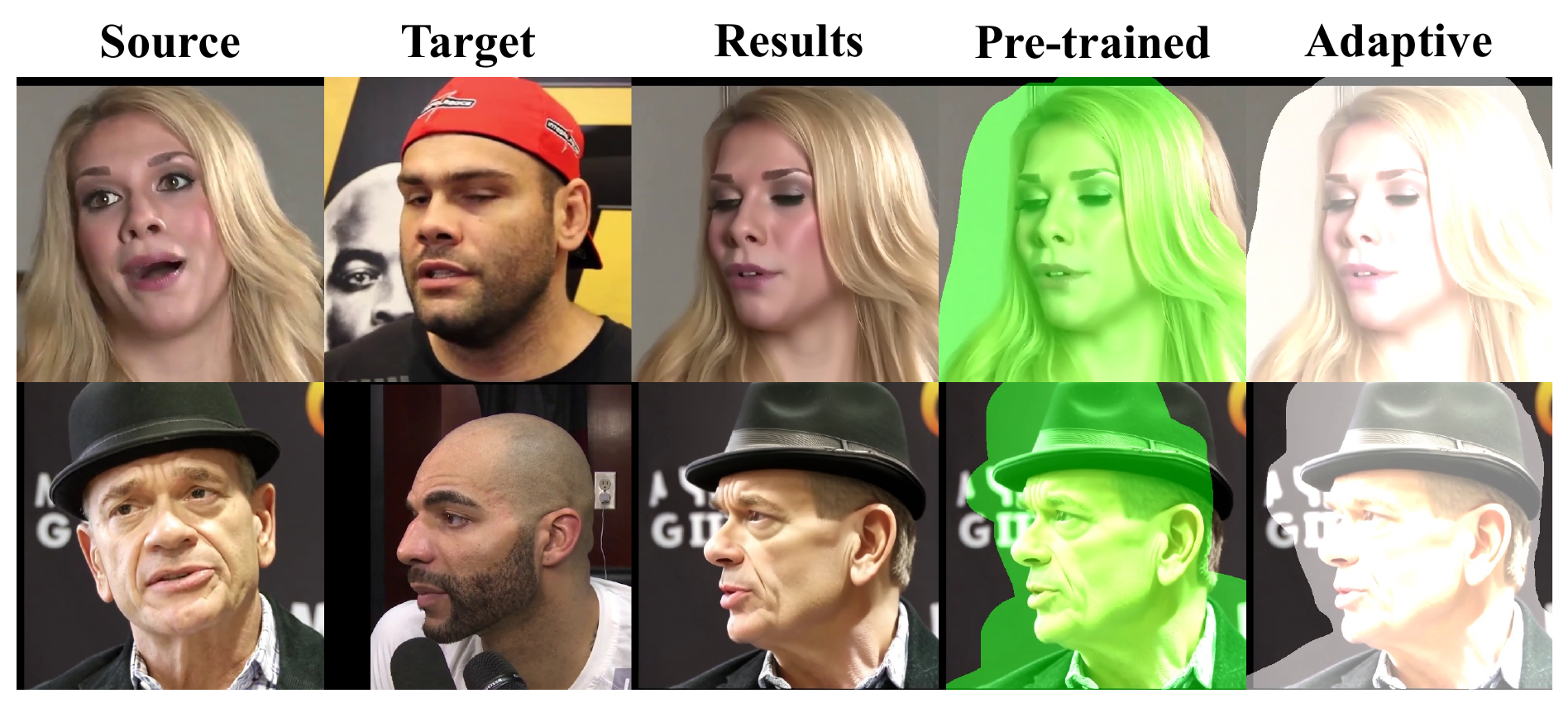}
    \caption{Comparisons with mask generated by pre-trained face parsing model (green) and $\varphi_{Re}$ (white). The green mask cannot fully cover the entire portrait. }
    \vspace{-1.7em}
    \label{fig:aap_vis}
\end{figure*}
It should be noted that we cannot directly utilize the pre-trained face parsing model in face reenactment. As shown in~\cref{fig:aap_vis} row 4, when the portrait area of the source image is larger (\eg, long hair and hat) than that in the target image, the green mask created by the pre-trained parsing model cannot fully cover the entire portrait and may result in artifacts at the boundary. However, the white mask created by $\varphi_{Re}$ in~\cref{fig:aap_vis} row 5 can encapsulate the whole portrait, as $\varphi_{Re}$ merely uses the source image and 3D landmarks as input, and exhibits excellent generalization when the source and target images possess different identities.

For face-swapping, the adapting area constitutes the facial region of the target image $I^T$. We employ a pre-trained face parsing model~\cite{bisenet} to predict the adapting area mask $M^{fg}_{Sw}$ of the target image $I^T$. Nonetheless, to accommodate face shape differences during testing, we designate the ground truth $M^{gt}_{Sw}$ as the region obtained by dilating the facial area outward.
\subsection{Identity Encoder}
\label{sec:ie}
As demonstrated by IP-Adapter-FaceID~\cite{ipadapter} and InstantID~\cite{instantid}, a high-level face embedding can ensure more robust identity preservation. As we observed, there is no need for heavy texture encoders~\cite{animateanyone} or additional identity networks~\cite{instantid} in face reenactment/swapping. By merely tuning a lightweight mapping module to map the face embedding into the fixed textual space, identity preservation is guaranteed. Specifically, given a face image $I^S$, the face embedding $f_{id}$ is obtained by a pre-trained face recognition model $E_{id}$~\cite{arcface}. Subsequently, a three-layer transformer decoder $\phi_{dec}$ is employed to project the face embedding  $f_{id}$ into the fixed text semantic space of the pre-trained diffusion model, obtaining the identity tokens. The specified number $N$ (we set $N=77$ in this paper) of learnable queries $q_{id}=\{q_1, q_2, \cdots, q_N\}$ in the transformer decoder constrains the sequence length of the identity embedding, ensuring it does not exceed the maximum length of the text embedding. Through this approach, the U-Net of the pre-trained diffusion model does not require any fine-tuning to adapt to the face embedding.

\subsection{Attribute Controller}
\label{sec:ac}

\noindent\textbf{Spatial Control.} 
In line with ControlNet~\cite{controlnet}, we create a copy of U-Net $\phi_{Ctl}$ and add spatial control $I_{Sp}$ as the conditioning input.
The spatial control image $I_{Sp}^S$/$I_{Sp}^T$ is obtained by combining the target motion landmarks $I^T_{lmk}$ and the non-adapting area obtained by the Adapting Area Predictor $\varphi_{Re}$ (or $\varphi_{Sw}$).
$$
I_{Sp}^S = I^S * (1-M^{fg}_{Re}) + I^T_{lmk},~\text{for face reenactment}, 
$$ 
$$
I_{Sp}^T = I^T * (1-M^{fg}_{Sw}) + I^T_{lmk},~\text{for face swapping}. 
$$
At this juncture, both reenactment and swapping tasks can be viewed as processes of performing conditional inpainting, utilizing the given identity and other missing attribute content, following the provided spatial control.

\noindent\textbf{Attribute Template.} 
Given identity and spatial control with part of the background, the attribute template is designed to supplement the missing information, including lighting and part of the background and hair. Attribute embeddings $f_{attr} \in \mathbb{R}^{257*d}$ are extracted from the attribute template ($I^S$ for reenactment and $I^T$ for swapping) using CLIP $E_{clip}$. To simultaneously obtain local and global features, we use both the patch tokens and the global token. The feature mapper module is also constructed as a three-layer transformer layer $\varphi_{dec}$ with learnable queries $q_{attr}=\{q_1, q_2, \cdots, q_K\}$, $K=77$.

\subsection{Strategies for Boosting Performance}

\noindent\textbf{Training.} 
\textit{\textbf{1)} Data Stream}: For both reenactment and face-swapping tasks, we use two images of the same person in different poses as source and target images. To support a ``one model for both task'' approach, we use a 50\% probability to choose between reenactment and face-swapping data streams during training, i.e., the spatial control and attribute template in the Attribute Controller use the data streams indicated by red and blue respectively. 
\textit{\textbf{2)} Condition Dropping for Classifier-free Guidance}: The conditions we need to drop include identity tokens and attribute tokens input into the U-Net and ControlNet cross-attention. We use a 5\% probability to simultaneously drop identity tokens and attribute conditions to enhance the realism of the image. To fully utilize the identity tokens for generation face images and improve identity preservation, we use an additional 45\% probability to drop attribute tokens.

\noindent\textbf{Inference.} 
\textit{\textbf{1)} Adapting Area Predictor
}: For reenactment, the input is the source (which is different from training) and corrected landmarks, and the output is the adapting area. For face-swapping, the input is the target, and the output is the adapting area. 
\textit{\textbf{2)} Negative Prompt for Classifier-Free Guidance}: For reenactment, negative prompts of both identity and attribute tokens are empty prompt embeddings. For face-swapping, to overcome the negative impact of the target identity in attribute tokens, we use the identity tokens of the target image as the negative prompt for identity tokens.

%% file: sections/4_experiments.tex
\section{Experiments}
\label{sec:4}
\subsection{Experimental Setup}




\noindent\textbf{Datasets.} 
During training, we leverage the VoxCeleb1 and VoxCeleb2~\cite{voxceleb2} dataset.
During the evaluation, we leverage the 491 test videos from the VoxCeleb1~\cite{voxceleb1} dataset and randomly sample 1,000 images in quantitative evaluation for face reenactment. We use FaceForensics++~\cite{faceforensics++} in quantitative evaluation for face swapping. 
We also spare 1,000 images from VoxCeleb2 for qualitative evaluation.
Following the preprocessing method in FOMM~\cite{fomm}, we crop faces from the original videos and resize them to 512$\times$512 for training and evaluation. 


\noindent\textbf{Evaluation Metrics.} 
For face reenactment, we use PSNR and LPIPS~\cite{lpips} to evaluate the reconstruction quality for same-identity reenactment. 
We use FID~\cite{fid} to evaluate the overall quality of the generated images. 
We use cosine similarity (CSIM) calculated by ~\cite{curricularface} to evaluate identity preservation. The motion transfer error is measured by Pose, Exp, and Gaze, which calculate the average Euclidean distances of pose, expression, and gaze coefficients between the generated and drive images. For face swapping, ID retrieval (ID) retrieves the closest face to evaluate identity modification, while Pose, Exp, and Gaze evaluate the attribute error between the generated faces and target faces.

%
%

\noindent\textbf{Implementation Details.} 
The Adapting Area Predictor is modified from the parsing model~\cite{bisenet}, with 6 input channels and 1 output channel. The identity-to-tokens is implemented with a 3-layer transformer decoder, a linear layer is added to project the identity feature dimensions to 768. The architecture of attribute-to-tokens is the same as the identity-to-tokens, except the input dimensions of the linear layer are consistent with the output dimensions of the CLIP model. We adopt the StableDiffusion v1-5~\cite{web:huggingface} as the pre-trained diffusion model and clip-vit-large-patch14~\cite{openclip} from OpenAI as the CLIP vision model in this paper. 
We train our face-adapter for 70,000 steps on 8$\times$V100 NVIDIA GPUs with a constant learning rate of 1e-4 and a batch size of 32. 
%

%

\subsection{Comparison with State-of-the-Art Methods}

\noindent\textbf{Face Reenactment.}
In~\cref{tab:reenact_quantitative_metric}, we compare with SoTA methods quantitatively on VoxCeleb1 test set, including GAN-based FOMM~\cite{fomm}, PIRenderer~\cite{pirenderer}, DG~\cite{dg}, TPSM~\cite{tpsm}, DAM~\cite{dam}, HyperReenact\cite{hyperreenact} and diffusion-based FADM~\cite{fadm}. FOMM, TPSM, and DAM are warping-based techniques, while PIRenderer and HyperReenact are 3DMM-based.

\begin{figure*}[tp]
    \centering
    \includegraphics[width=0.95\linewidth]{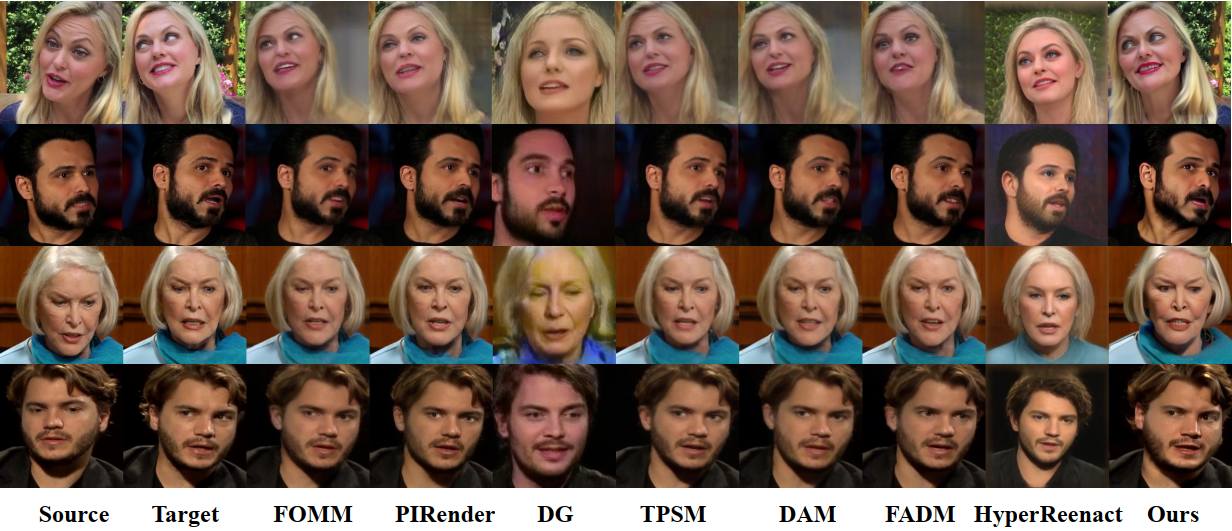}
    \caption{
        \textbf{Same-identity face reenactment results on Voxceleb2 test set.}
        Our method faithfully reconstructs the background and facial details.
    } 
    \label{fig:reenact_qualitative_comparison_same}
    \vspace{-1mm}
\end{figure*}
\begin{figure*}[]
    \centering
    \includegraphics[width=0.95\linewidth]{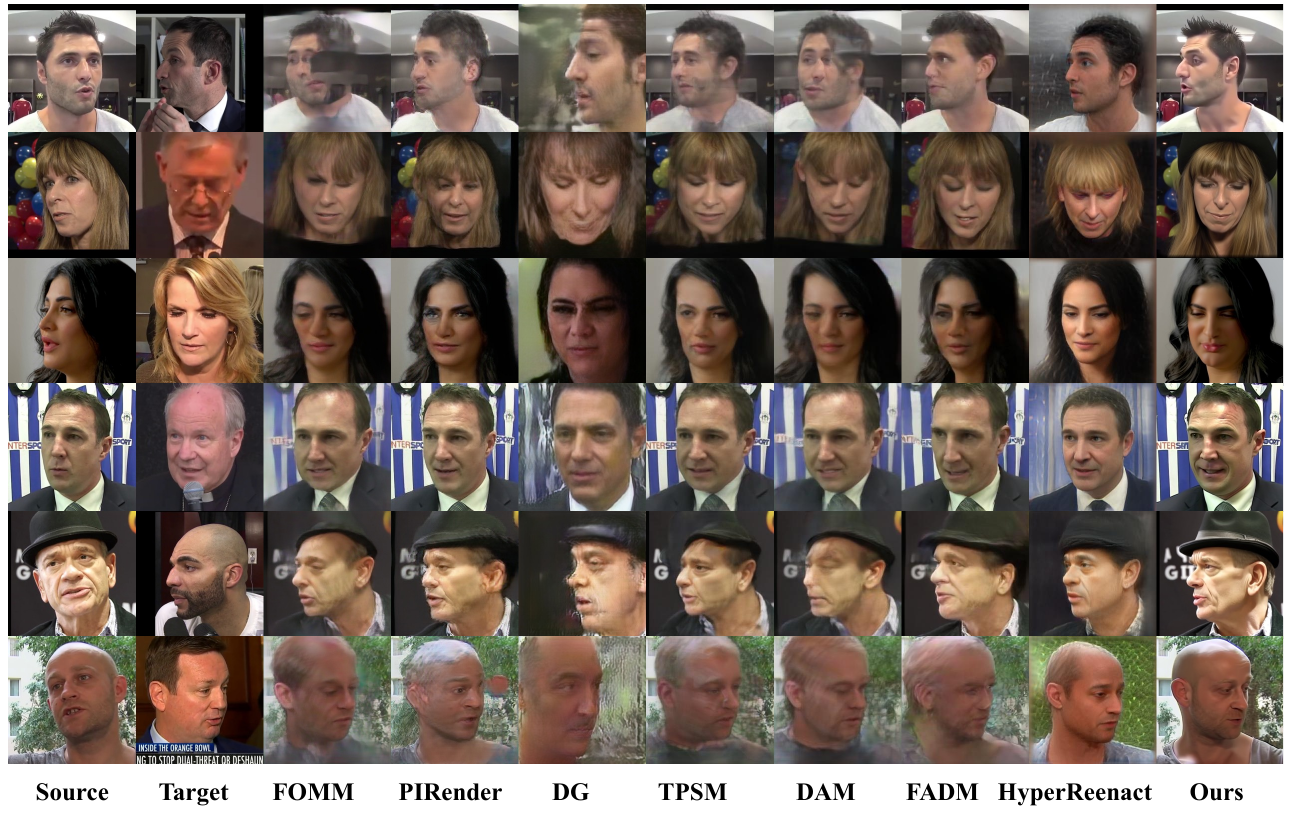}
    \vspace{-2mm}
    \caption{
        \textbf{Cross-identity face reenactment results on Voxceleb2 test set.}
        Our method significantly surpasses previous methods in terms of image quality and motion control accuracy, including pose, expression, and gaze, even under extreme poses. Moreover, we faithfully maintain consistency with the source in local details such as background and accessories, as well as global lighting. 
    } 
    \label{fig:reenact_qualitative_comparison_cross}
\end{figure*}

We achieve comparable or even optimal results in image quality. Owing to the Spatial Condition Generator, during training, incorporating the target background area in spatial condition avoids the interference of background motion. During inference, adding the source background in spatial condition significantly reduces the difficulty of generating backgrounds, improving background consistency. As a result, our method is capable of producing high-quality images with clear advantages in FID scores as well as in reconstruction metrics, \ie, PSNR and LPIPS.
In terms of motion control, our method performs well in pose and gaze error, but not as well in expression error.
As our landmarks are derived from D3DFR, both the reconstruction and projection processes, along with the sparsity of the landmarks, result in a loss of expression accuracy. Therefore, our method achieves a relatively moderate performance in terms of expression error.

In~\cref{fig:reenact_qualitative_comparison_same} and~\cref{fig:reenact_qualitative_comparison_cross}, we compare with SoTA methods qualitatively on VoxCeleb1 and Voxceleb2 test set.
The Spatial Condition Generator effectively ensures that our results are consistent with the source background and meanwhile reduces the training difficulty of the model, allowing it to focus more on face generation and improve the image quality. 
Freezing all parameters of the U-Net avoids overfitting and preserves as much of the powerful prior from the pre-trained diffusion model as possible. As a result, compared to other GAN-based methods and diffusion-based methods trained from scratch like FADM, our method is capable of generating faithful attribute details, \,  i.e., hair texture, hat, and accessories, that are consistent with the source image.
\begin{table*}[t]
\caption{Quantitative evaluations among SoTAs on Voxceleb1 test set. \textbf{Bold} and \underline{underline} correspond to the optimal and sub-optimal values, respectively.}
\setlength\tabcolsep{2.0pt}
\resizebox{\textwidth}{!}{%
\begin{tabular}{@{}lcccccccccccc@{}}
\toprule
\multirow{2}{*}{Methods} & \multicolumn{7}{c}{Same-Identity}                           & \multicolumn{5}{c}{Cross-Identity}      \\ \cmidrule(l){2-8}\cmidrule(l){9-13} 
                         & PSNR$\uparrow$ &LPIPS$\downarrow$&FID$\downarrow$& Exp$\downarrow$&Pose$\downarrow$&Gaze$\downarrow$&CSIM$ \uparrow$&Exp$\downarrow$&Pose$\downarrow$&Gaze$\downarrow$&CSIM$\uparrow$& FID$\downarrow$\\ \midrule
FOMM~\cite{fomm}         & 22.77          & \underline{0.1344}     & 31.19          & \underline{2.92}         & 0.0276         & \underline{0.0566}        & 0.8499         & 6.89          & 0.0644         & 0.1003                        & 0.539          & 51.57         \\
PIRenderer~\cite{pirenderer}     & 21.65          & 0.1388          & \underline{29.98}          & 3.08         & 0.0409         & 0.0798       & 0.819& \underline{6.42}& 0.0646         & 0.0963                        & 0.5361       & \textbf{40.71}         \\
DG~\cite{dg}& 14.01         & 0.4928  & 102.17          & 6.16         & 0.0707         & 0.112  & 0.0972  & 7.16          & 0.074         & 0.1287                       & 0.0834          & 102.61         \\
TPSM~\cite{tpsm}     & \underline{23.8}   & 0.1367          & 34.11 & \textbf{2.70}       & \textbf{0.0234} & 0.0627        & \textbf{0.8536}         & 6.58         & 0.0548& 0.0959                        & 0.5514          & 54.83         \\
DAM~\cite{dam}        & \textbf{23.85}  & 0.1484          & 38.6         & 2.87         & 0.027        & 0.0675        & \underline{0.8505 }        & 6.82          & 0.0636         & 0.1034                       & 0.5198          & 62.77        \\
HyperReenact~\cite{hyperreenact} &  15.73       &    0.3361       &   88.72       &  3.68     &   0.0381   & 0.0743   & 0.5455        & \textbf{5.94} & \underline{0.0452} & \underline{0.0812} & 0.4665 &     88.02  \\ \midrule
FADM~\cite{fadm}       & 22.70          & 0.1392          & 31.58          & 3.11         & 0.0324         & 0.086        & 0.8472         & 7.03          & 0.0786         & 0.1239                        & \underline{0.6152}          & 42.7        \\

Ours  &  22.36  &  \textbf{0.1281}  & \textbf{29.27}    &  3.24   & \underline{0.0243}  &  \textbf{0.0415}   &  0.7146       & 6.45 & \textbf{0.0355} & \textbf{0.0543} & \textbf{0.6429} &  \underline{41.09}     \\ \bottomrule

\end{tabular}%
}
\label{tab:reenact_quantitative_metric}
\end{table*}

\begin{figure*}[tp]
    \centering
    \includegraphics[width=1.0\linewidth]{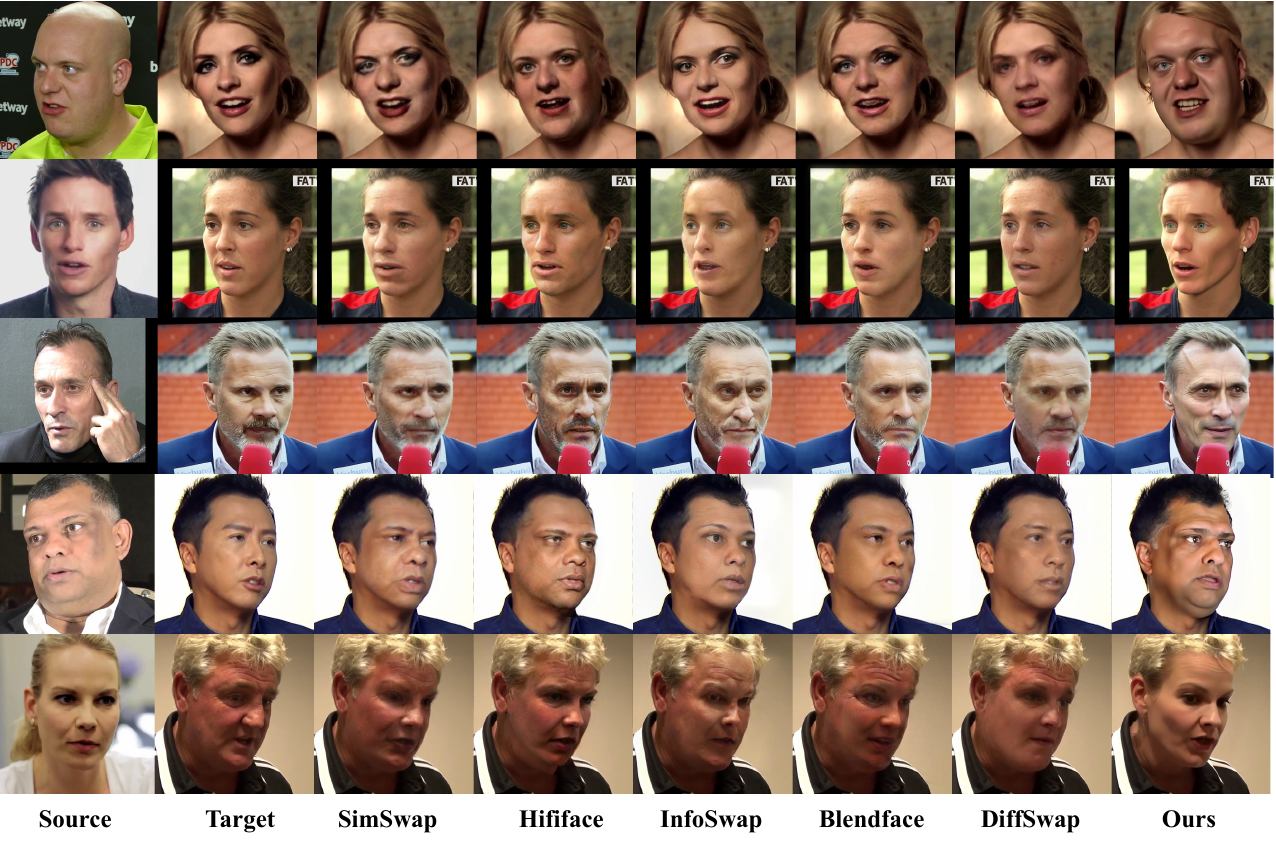}
    \caption{
        \textbf{Face swapping qualitative comparison results on Voxceleb2 test set.}
        Our method handles large facial shape changes effectively. It is capable of reasonably inpainting the blank area of the background caused by alternations in facial shape.
    } 
    \label{fig:swapping_qualitative_comparison_faceshape}
\end{figure*}

In addition to local details, the attribute tokens in the Attribute Controller effectively extract global illumination from the source image, significantly outperforming other methods. This further highlights the strengths and capabilities of our proposed approach in capturing both local and global features, leading to more realistic and accurate results.
Even when dealing with large poses, the Identity Encoder ensures robust identity preservation, and the pre-trained diffusion model reasonably generates attributes such as long hair that moves along with the face, demonstrating the superiority of our proposed adapter.

\begin{figure*}[tp]
    \centering
    \includegraphics[width=1.0\linewidth]{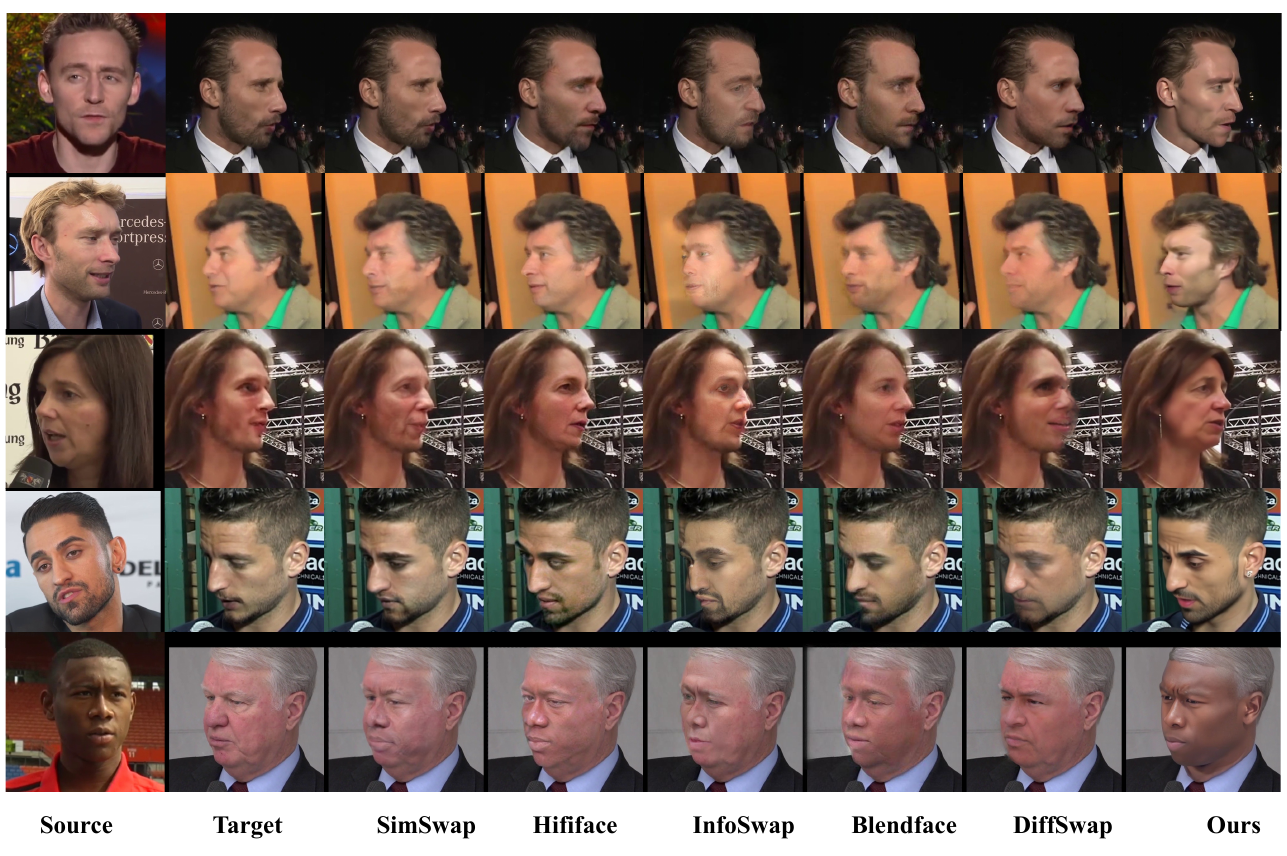}
    \caption{
        \textbf{Face swapping qualitative comparison results on Voxceleb2 test set.}
        Compared to previous methods, our approach faithfully maintains identity even under significant pose changes. 
    } 
    \label{fig:swapping_qualitative_comparison_largepose}
\end{figure*}

\noindent\textbf{Face Swapping.}
In~\cref{tab:swapping_quantitative_metrics}, we compare with SoTA methods quantitatively on FaceForensics++ test set, including  GAN-based FaceShifter~\cite{faceshifter}, SimSwap~\cite{simswap}, HifiFace~\cite{hififace}, InfoSwap~\cite{infoswap}, BlendFace~\cite{blendface} and diffusion-based DiffSwap~\cite{diffswap}.

Our 3D Landmark Projector helps to fuse the source face shape and target pose, expression and gaze to obtain the target motion landmarks in our spatial control. Our Adapting Area
The predictor allows ample space for changes in face shape while keeping enough background for inpainting. This combined spatial condition benefits the model's generation of natural images.
Although DiffSwap also utilizes shape-aware landmarks via D3DFR as spatial control, its inpainting process only takes place during DDIM sampling. Lacking a background reference makes it difficult for the model to generate clear facial results, which significantly affects image quality and ID similarity.
On the commonly used FaceForensics++ test set, our method is comparable to GAN-based methods in terms of ID, Pose, Exp and Gaze. 
Therefore, our method exhibits remarkable advantages in terms of ID while maintaining high motion accuracy compared to both GAN-based and diffusion-based SoTAs.


\cref{fig:swapping_qualitative_comparison_faceshape} and \cref{fig:swapping_qualitative_comparison_largepose} shows a qualitative comparison between our method and recent SoTA methods.
Previous methods struggle with handling significant changes in face shape and large pose. When transferring a thin-faced person to a fat-faced target image, these methods typically maintain the face shape of the target image, leading to a significant loss of identity. In contrast, our spatial control effectively addresses the issue of face shape changes. Unlike previous approaches that merely crop out the facial region, our Adapting Area Predictor allows ample space for changes in face shape. With the powerful generation capability of the pre-trained SD model, we can naturally complete the regions with facial shape variations. Furthermore, by using the identity tokens of the target image as a negative prompt during face-swapping inference, we further enhance the identity similarity with the source face.  As for large poses, previous methods struggle to generate plausible results, while our method directly generates faces from 3D landmarks without being affected by the pose.  

\begin{table*}[]
\centering
\caption{Quantitative results on the task of face swapping on FF++. Compared to the diffusion-based DiffSwap, our method significantly improves the metrics and achieves highly competitive results. \textbf{Note that our method can simultaneously perform both face reenactment and swapping}. \textbf{Bold} corresponds to the optimal values. $^{*}$: evaluated results are from the official code. \dag: evaluated results are from the officially released generated videos. }
\setlength\tabcolsep{16pt}
\resizebox{0.85\textwidth}{!}{%
\begin{tabular}{@{}llllll@{}}
\toprule
Methods      & ID $\uparrow$ & Pose$\downarrow$ & Exp$\downarrow$ & Gaze$\downarrow$ \\ \midrule
FaceShifter~\cite{faceshifter}\dag    &  87.99  & 0.0342     &  6.32   & 0.072          \\
SimSwap~\cite{simswap}$^{*}$     &  96.78  & \textbf{0.0261}     & \textbf{5.94}    &\textbf{0.0549} \\
HifiFace~\cite{hififace}\dag    & 94.26   & 0.0382    &  6.50   & 0.0573         \\
InfoSwap~\cite{infoswap}$^{*}$    & \textbf{99.26}  &  0.0371    &   7.25  &  0.0617         \\
BlendFace~\cite{blendface}$^{*}$    & 89.91   &  0.0286    &  6.15   &  0.0556         \\
\midrule
DiffSwap~\cite{diffswap}$^{*}$    & \textcolor{lightgray}{19.16}   & \textcolor{lightgray}{0.0237}    &  \textcolor{lightgray}{4.94} &  \textcolor{lightgray}{0.0665}      \\
Ours        & 96.47    &   0.0319   &   6.66  &  0.0607         \\ \bottomrule
\end{tabular}
}
\label{tab:swapping_quantitative_metrics}
\end{table*}

\subsection{Ablation Study and Further Analysis}

We conducted an ablation study on the Adapting Area Predictor and assessed the necessity of fine-tuning CLIP. For a fair comparison, all three models here were trained for 35,000 steps. Quantitative evaluations are conducted on Voxceleb1 cross-identity test set for both face reenactment and swapping tasks.

\begin{figure*}[htp]
    \centering
    \includegraphics[width=0.80\linewidth]{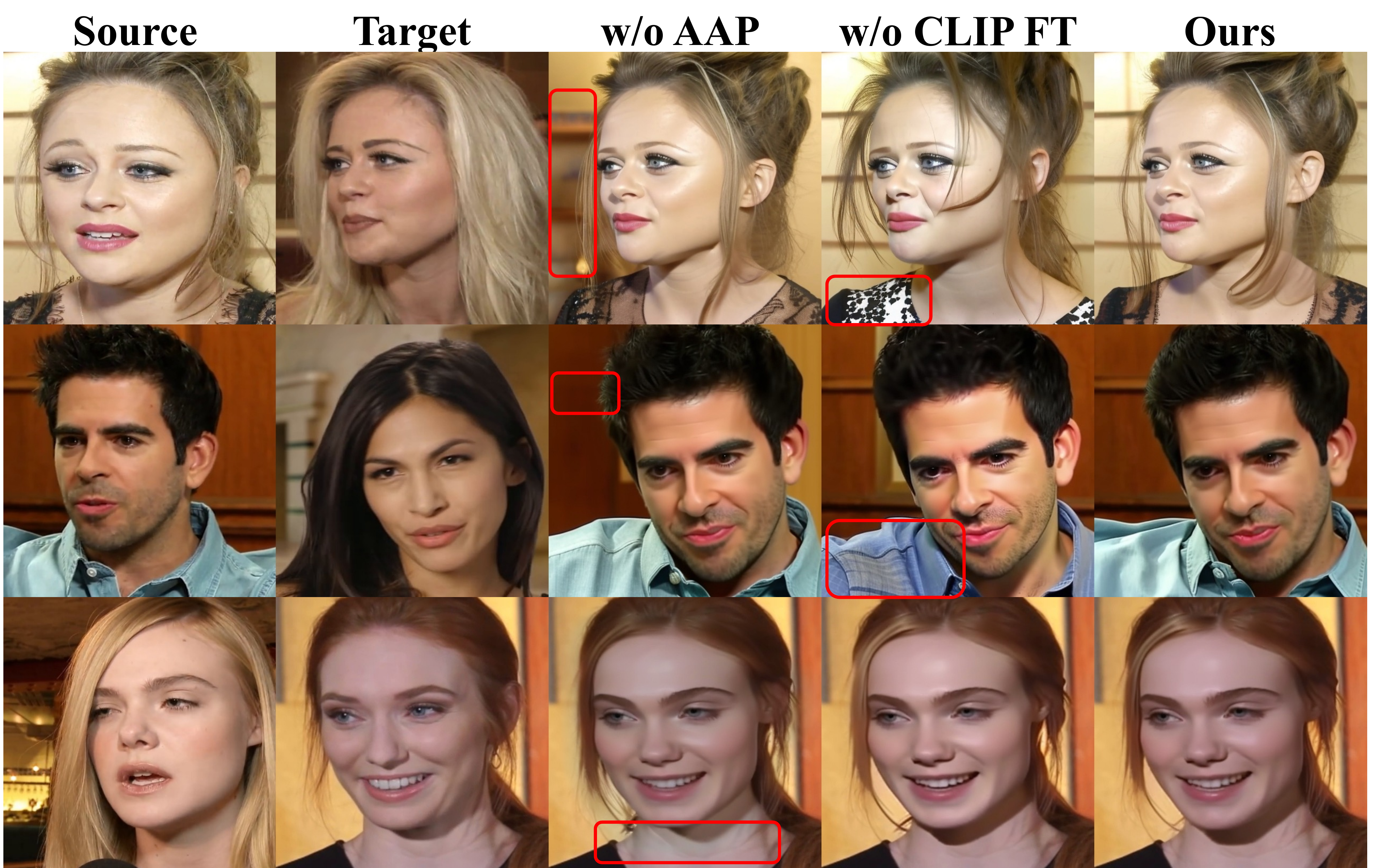}
    \caption{
        \textbf{Ablation study for Spatial Condition Generator and CLIP finetuning.} The red boxes highlight the artifacts in the picture. 
    } 
    \label{fig:ablation}
\end{figure*}

\noindent\textbf{Adapting Area Predictor.}
 As demonstrated in ~\cref{tab:ablation} and ~\cref{fig:ablation}, without the Adapting Area Predictor, the spatial control lacks the background and only includes landmarks from the 3D Landmark Projector. 
During training, the model extracts the background features from the source image in face reenactment, while using the target image background as the ground truth. This discrepancy tends to result in the model hallucinating background, and the model struggles to maintain consistency with the background of the source image during inference. As for face swapping, the model is not trained with inpainting task,  which leads to noticeable unnatural artifacts when blending the face with the surrounding area during inference.

\noindent\textbf{Fine-tuning CLIP for Extracting Attribute Features.}
As demonstrated in ~\cref{tab:ablation} and ~\cref{fig:ablation}, freezing the CLIP results in a decline in detailed attributes and image quality. The pre-trained CLIP is trained for discrimination tasks and lacks detailed texture features needed for generation tasks. Fine-tuning CLIP helps to extract detailed attribute features, including hair, clothing, part of the missing backgrounds, and global lighting; in addition to this, the fine-tuned CLIP model also extracts some features related to face identity, which benefits the identity similarity score in face reenactment.

\begin{table*}[htp]
\centering
\caption{Quantitative comparison of our model under different ablative configurations.}
\setlength\tabcolsep{8pt}
\resizebox{0.99\textwidth}{!}{%
\begin{tabular}{@{}lcccccccccc@{}}
\toprule
\multirow{2}{*}{Methods} & \multicolumn{5}{c}{Face Reenactmnet} & \multicolumn{5}{c}{Face Swapping} \\ \cmidrule(l){2-6} \cmidrule(l){7-11} 
& FID$\downarrow$ & Pose$\downarrow$  & Exp$\downarrow$ & Gaze$\downarrow$ & ID$\uparrow$ 
& FID$\downarrow$ & Pose$\downarrow$  & Exp$\downarrow$ & Gaze$\downarrow$ & ID$\uparrow$ \\ \midrule
w/o AAP      & 33.61  &  0.0281   & 3.72   &   0.045   & 0.6355    & 33.97   & 0.0395 & 6.13   & 0.0548    &  0.4530    \\
w/o CLIP FT  & 33.09  &  0.0287   & 3.74   &  0.0435   & 0.6474    & 31.97  & 0.0396 &  6.21  & 0.0540    &  0.4696    \\
Full Model   & 31.18  &  0.0266   & 3.61   &   0.0422  & 0.6616    &  30.78     & 0.0406 & 6.14   &  0.0547   &  0.4688    \\ \bottomrule
\end{tabular}
}
\label{tab:ablation}
\end{table*}

%% file: sections/5_conclusion.tex
\section{Conclusion}
In this paper, we present a novel Face-Adapter framework, a plug-and-play facial editing adapter that supports fine control over identity and attributes for pretrained diffusion models. Utilizing only one model, this adapter effectively addresses face reenactment and swapping tasks, surpassing previous state-of-the-art GAN-based and diffusion-based methods.
It comprises a Spatial Condition Generator, an Identity Encoder, and an Attribute Controller. 
The Spatial Condition Generator is used to predict the 3D prior landmarks and the mask of the area that needs to be changed, working with the Identity Encoder and Attribute Controller to formulate reenactment and swapping tasks as conditional inpainting with sufficient spatial guidance, identity, and essential attributes. Through reasonable and highly decoupled condition design, we unleash the generative capabilities of pretrained diffusion models for face reenactment and swapping tasks. 
Extensive qualitative and quantitative experiments demonstrate the superiority of our method. 

\vspace{3mm}
\noindent\textbf{Limitations} 
Our unified model is unable to achieve temporal stability in video face reenactment/ swapping, which requires incorporating additional temporal fine-tuning in the future.

\noindent\textbf{Potential Social Impact} 
For the first time, we explore a lightweight framework based on diffusion for simultaneous face reenactment and swapping, which has higher practical value while improving the quality of generated content. However, the potential misuse of Face-Adapter can lead to privacy invasion, misinformation spread, and ethical concerns. 
To mitigate these risks, both visible and invisible digital watermarks can be incorporated to help identify the origin and authenticity of the content.
On the other side, Face-Adapter can contribute to the field of forgery detection, further enhancing the ability to identify and combat deepfakes.